\documentclass{article}

\PassOptionsToPackage{numbers, compress}{natbib}
\usepackage[preprint]{main}
\usepackage[utf8]{inputenc} 
\usepackage[T1]{fontenc}   
\usepackage{hyperref}       
\usepackage{url}            
\usepackage{booktabs}       
\usepackage{amsfonts}       
\usepackage{nicefrac}       
\usepackage{microtype}      
\usepackage{xcolor}         

\usepackage{array} 
\usepackage{graphicx}
\usepackage{amsmath}
\usepackage{amssymb}
\usepackage{arydshln} 
\usepackage{amsfonts}
\usepackage{epsfig}
\usepackage{algorithm}
\usepackage{algpseudocode}
\usepackage{booktabs}
\usepackage{tabularx}   
\usepackage{mathrsfs}
\usepackage{wrapfig}
\usepackage{booktabs}
\usepackage{color}
\usepackage{textcomp}
\usepackage{bbding}
\usepackage{pifont}
\usepackage{wasysym}
\usepackage{subcaption}
\usepackage{xcolor,colortbl}
\usepackage{overpic}
\usepackage{microtype}
\usepackage{multirow}
\usepackage{makecell}
\usepackage{hhline}
\usepackage[american]{babel}
\usepackage{ragged2e}
\usepackage{paralist}
\definecolor{lightgray}{gray}{.94}
\definecolor{tinygray}{gray}{.96}

\newcommand{\eg}{\textit{e}.\textit{g}.}

\usepackage{multirow}
\usepackage{epsfig}
\usepackage{adjustbox}

\title{PolarVLM: Bridging the Semantic-Physical Gap in Vision-Language Models}

\author{%
  Yuliang Li$^{1\#}$ \quad
  Chu Zhou$^{2\#}$ \quad
  Heng Guo$^{1*}$ \quad
  Boxin Shi$^{3}$ \quad
  Imari Sato$^{2,4}$ \quad
  Zhanyu Ma$^{1}$ \\
  $^{1}$Beijing University of Posts and Telecommunications, China \\
  $^{2}$National Institute of Informatics, Japan \\
  $^{3}$Peking University, China \\
  $^{4}$The University of Tokyo, Japan \\
  \texttt{\{liyuliang, guoheng, mazhanyu\}@bupt.edu.cn} \\
  \texttt{zhou\_chu@hotmail.com, shiboxin@pku.edu.cn, imarik@nii.ac.jp}
}

\begin{document}

\maketitle

\begingroup
\renewcommand{\thefootnote}{\#}
\footnotetext[0]{Equal contribution.}
\renewcommand{\thefootnote}{*}
\footnotetext[0]{Corresponding author.}
\endgroup

\begin{abstract}
 Mainstream vision-language models (VLMs) fundamentally struggle with severe optical ambiguities, such as reflections and transparent objects, due to the inherent limitations of standard RGB inputs. While polarization imaging captures polarimetric physical parameters that resolve these ambiguities, existing methods are constrained by fixed-format outputs and remain isolated from open-ended reasoning. To bridge this semantic-physical gap, we introduce \textbf{PolarVLM}, the first multimodal framework integrating polarimetric physical parameters into VLMs. By employing a dual-stream architecture and a progressive two-stage training strategy, PolarVLM effectively prevents physical misinterpretations while preserving general visual abilities. Complementing our architecture, we construct \textbf{PolarVQA}, the first benchmark for polarization-aware VQA, featuring 75K physics-grounded instruction-tuning pairs targeting reflective and transparent scenes. Experiments show that PolarVLM surpasses the RGB baseline by 25.4\% overall across five evaluation tasks, with remarkable gains of 26.6\% in reflection recognition and 34.0\% in glass counting, successfully unlocking physics-aware semantic understanding.
\end{abstract}

\section{Introduction}
\label{sec:intro}

\begin{figure*}[t]
  \centering
  \includegraphics[width=\linewidth]{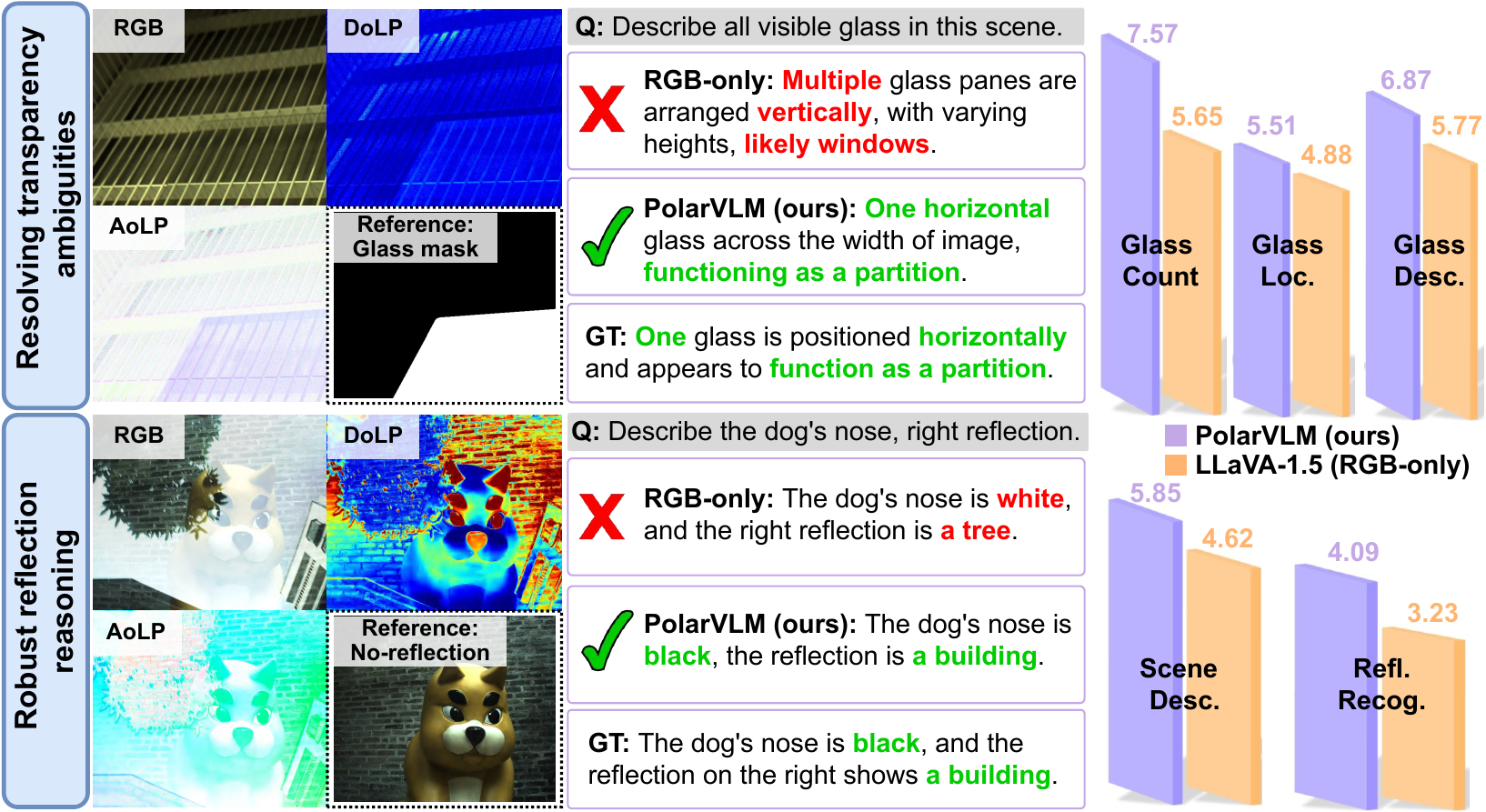}
  \caption{
    PolarVLM overcomes optical ambiguities through physics-aware multimodal reasoning. By integrating polarimetric physical parameters, PolarVLM significantly outperforms conventional RGB-only baselines (\eg, fine-tuned LLaVA-1.5~\cite{liu2024improved}) in resolving transparency ambiguities and performing robust reflection reasoning, both qualitatively and quantitatively.
  }
  \label{fig:teaser_new}
\end{figure*}

Mainstream vision-language models (VLMs) excel at open-ended visual question answering (VQA) using RGB inputs~\cite{liu2023visual,li2023blip,liu2024improved}. However, real-world environments frequently feature specular reflections and transparent objects, which pose systematic challenges to this paradigm. These phenomena introduce inherent \textit{optical ambiguities}: because standard RGB sensors capture only intensity and color, they cannot physically disentangle actual objects from reflected virtual content, nor perceive transparent surfaces lacking distinct visual boundaries. Relying exclusively on such ambiguous photometric features, existing VLMs are highly susceptible to severe perceptual illusions. Consequently, as illustrated by the RGB-only failures in Fig.~\ref{fig:teaser_new}, even state-of-the-art models fail to ground accurate semantic reasoning in these common physical scenarios. 

Polarization imaging breaks this bottleneck by capturing polarimetric physical parameters (\eg, Degree of Linear Polarization (DoLP) and Angle of Linear Polarization (AoLP)) that remain invisible to standard RGB sensors~\cite{goldstein2017polarized,collett2005field}. While these parameters have proven effective for low-level vision tasks like reflection removal~\cite{lyu2019reflection,wang2025polarized,yao2025polarfree} and transparent object segmentation~\cite{mei2022glass,shao2023transparent}, their potential for high-level semantic reasoning remains largely untapped. This is because existing approaches are strictly constrained by fixed-format outputs, fundamentally preventing them from supporting open-ended dialogue. Moreover, a critical void of polarization-text instruction data leaves modern VLMs entirely blind to these physical parameters. Therefore, there is a compelling need to develop a unified multimodal framework and benchmark that jointly integrate RGB and polarimetric cues, injecting physics-aware perception into open-ended VQA to unlock accurate semantic understanding.

Unlocking this capability, however, presents non-trivial challenges spanning both model architecture and data alignment. On the architectural front, directly injecting polarimetric physical parameters into standard VLMs leads to a severe representation gap. The physical properties encoded by these parameters differ from the appearance-based semantic features expected by pretrained visual encoders. A naive workaround is a cascaded approach that utilizes polarization solely for low-level image preprocessing. However, this strategy creates an opaque pipeline where early estimation errors inevitably accumulate. Consequently, the language model remains isolated from the raw physical signals, precluding genuine physics-aware reasoning. It is therefore critical to design a dedicated fusion mechanism that preserves the distinct strengths of both modalities. Concurrently, on the data front, existing polarimetric datasets offer only pixel-level annotations without language descriptions. The challenge lies in bridging this semantic-physical gap by translating dense physical measurements into high-fidelity textual supervision. Crucially, these annotations must strictly demonstrate the boundary of what polarization can actually express (\eg, geometric structure and material properties). This precision is essential to prevent the model from misinterpreting physical parameters as ordinary visual textures, or erroneously associating them with unrelated semantic concepts.

To address these limitations, we introduce \textbf{PolarVLM}, the first multimodal framework specifically designed to bridge the semantic-physical gap in open-ended VQA. By anchoring the language model in polarimetric physics, PolarVLM effectively resolves severe optical ambiguities for reflections and transparent objects (see Fig.~\ref{fig:teaser_new}). Overcoming the isolation of cascaded approaches, we construct a dual-stream architecture upon LLaVA-1.5~\cite{liu2024improved}. This design injects physical measurements via a dedicated polarization branch while robustly preserving pretrained RGB semantic priors. The resulting heterogeneous features are integrated via token-level concatenation, empowering the language model's self-attention layers to dynamically cross-reference physics-aware and appearance-based tokens. To prevent the model from misinterpreting physical parameters as ordinary visual textures, we propose a progressive two-stage training strategy. By initially aligning polarimetric features to the semantic space prior to dual-stream joint instruction tuning, we ensure that physical signals correctly ground the model's reasoning without disrupting its inherent visual-linguistic capabilities.

Complementing our architectural design, we introduce \textbf{PolarVQA} to bridge the critical void in polarimetric instruction data. We formulate a physics-guided data generation pipeline that integrates visual foundation models~\cite{xiao2024florence}, exact polarimetric computations, and large language models~\cite{hurst2024gpt}. This pipeline successfully translates low-level physical measurements into high-fidelity textual supervision while strictly preventing the introduction of unrelated semantic concepts. Built upon raw multimodal captures from public datasets~\cite{yao2025polarfree,mei2022glass}, we construct 28.5K physics-grounded captions and 46.8K diverse instruction-tuning pairs. Structurally, the benchmark is unified under two core dimensions: \textit{resolving transparency ambiguities} and \textit{robust reflection reasoning}, which feature a dedicated \textit{counterfactual reasoning} subset explicitly designed to penalize reflection-induced object hallucinations. By addressing these severe optical ambiguities, PolarVQA establishes the first comprehensive benchmark for training and evaluating physics-aware vision-language models.

In summary, our primary contributions include:
\begin{compactitem}
    \item \textbf{PolarVLM}, the first multimodal framework integrating polarimetric physical parameters into VLMs for open-ended VQA under severe optical ambiguities.
    \item \textbf{A dual-stream architecture and progressive training strategy} to bridge the semantic-physical gap, preventing physical misinterpretations while preserving general visual abilities.
    \item \textbf{PolarVQA}, the first benchmark for polarization-aware VQA, providing 75K physics-grounded captions and instruction-tuning pairs targeting reflective and transparent scenes.
\end{compactitem}

\section{Related work}
\label{sec:related_work} 

\noindent\textbf{Polarization-based vision.} 
Since the polarization properties of light are fundamentally governed by surface normals, textures, and material characteristics, polarization has become an indispensable physical cue for overcoming the limitations of conventional appearance-based approaches. In the realm of 3D vision, polarization has been widely exploited to solve a variety of ill-posed problems, including shape and normal estimation~\cite{deschaintre2021deep, ba2020deep, lyu2023shape, lei2022shape, cao2023multi}, 3D reconstruction and inverse rendering~\cite{han2024nersp, chen2024pisr, dave2022pandora}, and depth sensing~\cite{kadambi2017depth, tian2023dps}. Furthermore, because reflection and scattering consistently alter the polarization state of light, polarimetric cues have been extensively applied to low-level vision tasks such as reflection removal~\cite{lyu2019reflection, wang2025polarized, yao2025polarfree, lyu2022physics, lei2020polarized}, dehazing~\cite{schechner2001instant, zhou2021learning}, high dynamic range imaging~\cite{wu2020hdr, zhou2023polarizationHDR}, color constancy~\cite{ono2022degree}, and shadow removal~\cite{zhou2025polarization}. Most critically for complex real-world scenes, polarization provides unique discriminative features for transparent object perception~\cite{mei2022glass, shao2023transparent, kalra2020deep}, scenarios where purely RGB-based methods suffer from severe generalization bottlenecks due to the absence of explicit physical constraints~\cite{zhang2018single, wan2019corrn}. Despite their significant progress in these respective domains, existing methods remain confined to task-specific modules with fixed-format outputs, while PolarVLM pioneers the translation of polarimetric physical quantities into language supervision to enable open-ended, physics-grounded visual question answering.

\noindent\textbf{Vision-language models.}
Recent VLMs have successfully aligned pretrained visual encoders with large language models. Pioneer frameworks like LLaVA~\cite{liu2023visual, liu2024improved} and MiniGPT-4~\cite{zhu2023minigpt} project visual tokens via multi-layer perceptrons, while BLIP-2~\cite{li2023blip} and InstructBLIP~\cite{dai2023instructblip} employ learnable queries and cross-attention mechanisms. More recently, Qwen2-VL~\cite{wang2024qwen2} and InternVL~\cite{chen2024internvl} have introduced dynamic resolution and scaled language backbones. However, these general-purpose VLMs rely merely on RGB inputs. Their perceptual failures in reflective and transparent scenarios stem not from insufficient model capacity, but from the lack of physical information within the RGB modality itself.

\noindent\textbf{Incorporating non-RGB modalities.}
To transcend the physical limitations of RGB, researchers have increasingly explored extending vision-language alignment to diverse modalities. Foundational efforts successfully extended Contrastive Language-Image Pre-training (CLIP)~\cite{radford2021learning} to point clouds~\cite{zhang2022pointclip}, video~\cite{xu2021videoclip}, and audio~\cite{guzhov2022audioclip}, while unified frameworks like ImageBind~\cite{girdhar2023imagebind} have established shared embedding spaces across multiple sensory inputs. Furthermore, domain-specific sensors are progressively being integrated into VLMs to tackle extreme conditions; notable examples include tactile sensing for physical deformation~\cite{yang2024binding}, event cameras for high-dynamic scenes~\cite{li2025eventvl}, and thermal imaging for low-light environments~\cite{jiang2024infrared}. Despite successes with other modalities, incorporating polarization into VLMs is hindered by its complex information overlap with RGB and a critical lack of text-aligned datasets. PolarVLM resolves these challenges, introducing the first dual-stream framework and companion PolarVQA benchmark for physics-grounded open-ended reasoning.

\section{The PolarVLM framework}
\label{sec:framework}

\subsection{Physical image formation and polarimetric encoding}
To introduce physical perception into vision-language models, it is essential to first bridge the gap between raw optical measurements and neural representations. Standard polarization cameras employ a micro-polarizer array to capture the modulation of light. According to Malus's law~\cite{collett2005field}, the measured intensity $I_{\alpha}$ passing through a linear polarizer oriented at angle $\alpha$ can be expressed in terms of the Stokes parameters $\mathbf{S} = [S_0, S_1, S_2]^{\top}$:
\begin{equation}
    I_{\alpha} = \frac{1}{2} S_0 + \frac{1}{2} S_1 \cos(2\alpha) + \frac{1}{2} S_2 \sin(2\alpha),
\label{eq:malus}
\end{equation}
where $S_0$ denotes the total light intensity, and $S_1$ and $S_2$ characterize the linearly polarized components. In a single exposure, the sensor captures four spatially aligned directional intensities: $I_{0^\circ}$, $I_{45^\circ}$, $I_{90^\circ}$, and $I_{135^\circ}$. Based on Eq.~\ref{eq:malus}, the Stokes parameters can be strictly decoupled as:
\begin{equation}
    S_0 = I_{0^\circ} + I_{90^\circ}, \quad S_1 = I_{0^\circ} - I_{90^\circ}, \quad S_2 = I_{45^\circ} - I_{135^\circ}.
\end{equation}

Once the Stokes vector is obtained, the fundamental physical properties of the incident light, namely, the Degree of Linear Polarization (DoLP, denoted as $P$) and the Angle of Linear Polarization (AoLP, denoted as $\Phi$), can be analytically derived:
\begin{equation}
    P = \frac{\sqrt{S_1^2 + S_2^2}}{S_0}, \quad \Phi = \frac{1}{2}\arctan\left(\frac{S_2}{S_1}\right).
\end{equation}
Here, $P \in [0,1]$ indicates the proportion of linearly polarized light, serving as a strong physical cue for separating reflections and revealing transparent surfaces. $\Phi \in [0, \pi)$ represents the oscillation orientation of the electric field. However, directly feeding $\Phi$ into a pretrained visual encoder introduces severe domain mismatch issues. Because the physical angle is $\pi$-periodic, values near $0$ and $\pi$ represent nearly identical physical states but exhibit maximal numerical distance. This angular discontinuity disrupts the continuous feature space expected by standard convolutional or transformer-based encoders.

To seamlessly inject these physical properties into the vision-language domain, we design a continuous, neural-friendly polarimetric representation. We map the discontinuous $\Phi$ into a continuous 2D Cartesian space via trigonometric transformations $\sin(2\Phi)$ and $\cos(2\Phi)$. Notably, this is physically equivalent to utilizing the normalized Stokes parameters $S_2 / \sqrt{S_1^2 + S_2^2}$ and $S_1 / \sqrt{S_1^2 + S_2^2}$. By concatenating these components with the DoLP, we construct a normalized polarization input:
\begin{equation}
    \mathbf{X}_{\text{pol}} = [P,\; \sin(2\Phi),\; \cos(2\Phi)].
\end{equation}
This physics-grounded encoding strictly preserves the discriminative power of the original polarimetric measurements while eliminating periodic boundary ambiguities, explicitly aligning the physical signal distribution with the input requirements of the subsequent dual-stream architecture.

\begin{figure*}[t]
  \centering
  \includegraphics[width=\linewidth]{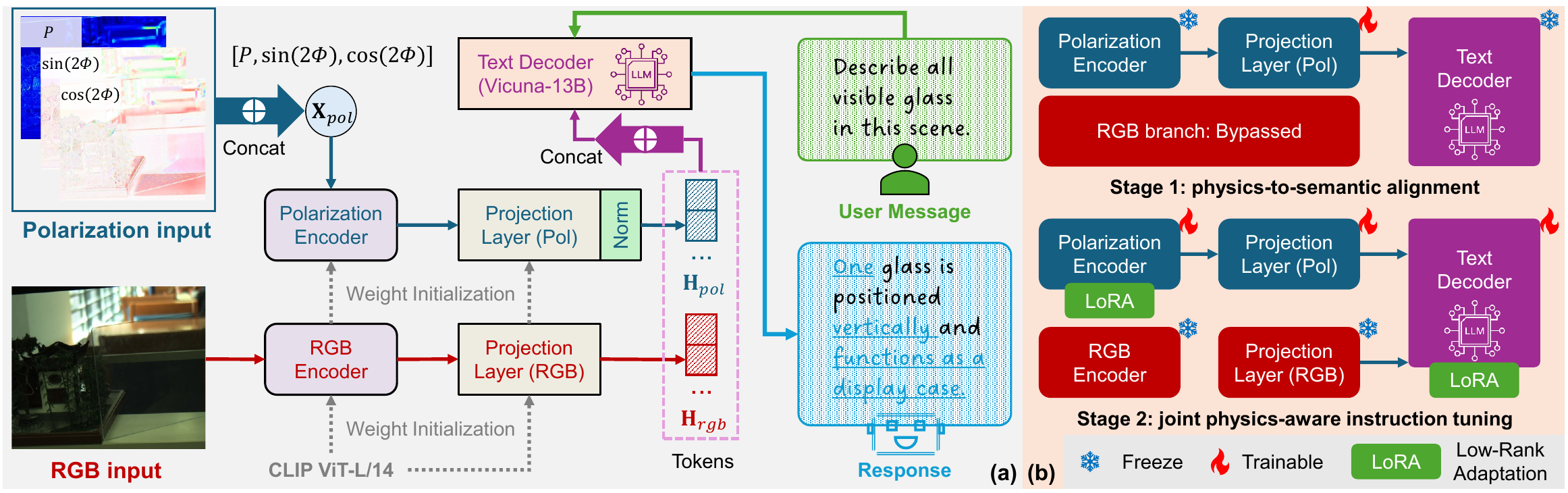}
  \caption{
   Overall design of PolarVLM. (a) Dual-stream physics-aware architecture with sequence-concatenation modality fusion. (b) Progressive two-stage training strategy to avoid modality collapse and achieve robust physical-semantic alignment.
  }
  \label{fig:architecture}
\end{figure*}

\subsection{Dual-stream architecture and modality fusion}
The core philosophy behind PolarVLM is to inject polarimetric physical perception into the language model without compromising its foundational semantic reasoning capabilities. Directly modifying the visual encoder to accept additional input channels would inevitably disrupt the pretrained semantic priors. Therefore, as shown in Fig.~\ref{fig:architecture}, we design a structurally symmetric dual-stream architecture built upon the LLaVA-1.5~\cite{liu2024improved} framework, explicitly decoupling appearance processing from physical feature extraction. The RGB encoder and its projection layer remain completely frozen throughout training, providing an anchor for stable semantic understanding.

\noindent\textbf{Physics-aware visual branch.}
To extract representations from the polarimetric input $\mathbf{X}_{\text{pol}}$, we construct a parallel visual branch adopting the same CLIP ViT-L/14 architecture~\cite{radford2021learning} as the RGB pathway. Rather than training this branch from scratch, we initialize it with pretrained RGB weights. This design choice is motivated by the strong morphological affinities between polarimetric parameters and low-level visual features: DoLP resembles an edge map~\cite{zhou2025learning, zhou2025pidsr}, while the spatial gradients of the trigonometric AoLP channels closely mimic a surface normal map~\cite{peters2023pcon, zhou2025polarimetric}. Consequently, the edge detection and texture extraction priors inherently learned by the pretrained vision model can effectively transfer to the physical domain. The extracted physical features are then mapped into the LLM embedding space via an MLP projection layer. Crucially, because the numerical distributions of physical and appearance features differ systematically, we introduce a normalization layer at the output of the polarization projector. This aligns the latent magnitudes of the two modalities, preventing the subsequent self-attention mechanism from being dominated by either modality due to scale imbalance.

\noindent\textbf{LLM-delegated modality fusion.}
Given the two heterogeneous sets of visual tokens, $\mathbf{H}_{\text{rgb}}$ and $\mathbf{H}_{\text{pol}}$, an appropriate fusion mechanism is critical. Conventional fusion strategies exhibit distinct limitations in this multimodal context: residual addition implicitly assumes strict semantic additivity at corresponding spatial locations, which fundamentally violates the orthogonal physical meanings of RGB and polarization. Conversely, introducing auxiliary cross-attention modules yields excessive learnable parameters, risking severe overfitting when fine-tuning on limited multimodal data. To circumvent these issues, PolarVLM adopts sequence concatenation. By simply concatenating $\mathbf{H}_{\text{rgb}}$ and $\mathbf{H}_{\text{pol}}$ along the sequence dimension, we impose no restrictive prior assumptions. Instead, we elegantly delegate the cross-modal information routing entirely to the robust self-attention mechanism of the LLM. This allows the model to dynamically cross-reference physics-aware tokens specifically when the appearance-based tokens are corrupted by specular reflections or transparency.

\section{Progressive two-stage training strategy}
\label{sec:training_strategy}

Training a heterogeneous multimodal framework from scratch is notoriously prone to modality collapse, where the language model relies entirely on familiar RGB representations and effectively ignores newly introduced physical signals. To avoid this and ensure robust physical grounding, PolarVLM employs a progressive two-stage training paradigm.

\noindent\textbf{Stage 1: physics-to-semantic alignment.}
The primary objective of this stage is to strictly enforce the semantic grounding of the polarimetric signals. As illustrated in Fig.~\ref{fig:architecture}, we explicitly bypass the RGB branch and feed only the physics-aware tokens $\mathbf{H}_{\text{pol}}$ into the language model. This unimodal bottleneck design forces the polarization branch to establish an independent semantic representation, categorically preventing the model from developing a lazy reliance on RGB priors during early training. Architecturally, we freeze the LLM and the polarization encoder's transformer blocks, updating only the polarization patch embedding layer and the MLP projector. This design is highly deliberate: the patch embedding directly ingests the normalized physical values and thus requires domain adaptation to map continuous physical distributions into discrete semantic tokens, whereas the higher-level spatial reasoning capabilities of the ViT are domain-agnostic and should be preserved.

\noindent\textbf{Stage 2: joint physics-aware instruction tuning.}
Having established a foundational semantic understanding of polarization, the model transitions to dual-stream instruction tuning, as shown in Fig.~\ref{fig:architecture}. Here, both modalities are activated, and the concatenated visual tokens $[\mathbf{H}_{\text{rgb}}, \mathbf{H}_{\text{pol}}]$ are jointly processed by the language model. The network is trained on complex, multi-turn VQA tasks involving reflections and transparency. To endow the model with advanced reasoning while preventing catastrophic forgetting, we freeze the polarization patch embedding and instead adapt the polarization ViT's attention modules alongside the LLM via Low-Rank Adaptation (LoRA)~\cite{hu2022lora}. This creates a decoupled adaptation scheme: Stage 1 achieves input-level physical alignment, while Stage 2 realizes high-level cross-modal semantic adaptation. 

\noindent\textbf{Training objective.}
Throughout both stages, we optimize the model using the standard autoregressive next-token prediction objective. Let $\mathbf{H}$ denote the visual tokens provided to the language model, where $\mathbf{H} = \mathbf{H}_{\text{pol}}$ in Stage 1, and $\mathbf{H} = [\mathbf{H}_{\text{rgb}}, \mathbf{H}_{\text{pol}}]$ in Stage 2. Given the embedded instruction tokens $\mathbf{H}_{\text{instruct}}$, the model maximizes the likelihood of generating the target answer sequence:
\begin{equation}
    \mathcal{L}(\theta) = - \sum_{i=1}^{L} \log p_\theta(y_i \mid \mathbf{H}, \mathbf{H}_{\text{instruct}}, y_{<i}), \quad y = \{y_1, \dots, y_L\},
\end{equation}
where $\theta$ denotes the trainable parameters of the respective stage, $L$ is the sequence length, and $y_{<i}$ represents the previously generated tokens.

\section{The PolarVQA benchmark}
\label{sec:benchmark}

To effectively train PolarVLM and rigorously evaluate its physics-aware reasoning capabilities, we construct PolarVQA, the first comprehensive vision-language benchmark targeting specular reflections and transparent objects. We source raw multi-angle polarimetric measurements and RGB images from two public datasets, PolarFree~\cite{yao2025polarfree} and RGBP-Glass~\cite{mei2022glass}. However, since these datasets provide only pixel-level physical annotations without natural language descriptions, they cannot be directly utilized for vision-language instruction tuning.

\noindent\textbf{Physics-guided data generation.}
Rather than relying on manual labeling, we design an automated, physics-guided data generation pipeline to translate raw physical measurements into high-fidelity textual supervision. At a high level, this pipeline adopts a visual foundation model (Florence-2-large~\cite{xiao2024florence}) for spatial structural extraction, precise polarimetric computations for physical grounding, and a large language model (GPT-4o-mini~\cite{hurst2024gpt}) for final text synthesis. To prevent the language model from hallucinating and to avoid introducing spurious supervisory signals into the polarization modality, we explicitly strip RGB-specific semantics (\eg, color and texture) from the polarization prompts. The detailed architecture of this generation pipeline is provided in Appendix~\ref{supp:pipeline}.

\begin{figure*}[t]
    \centering
    \includegraphics[width=\textwidth]{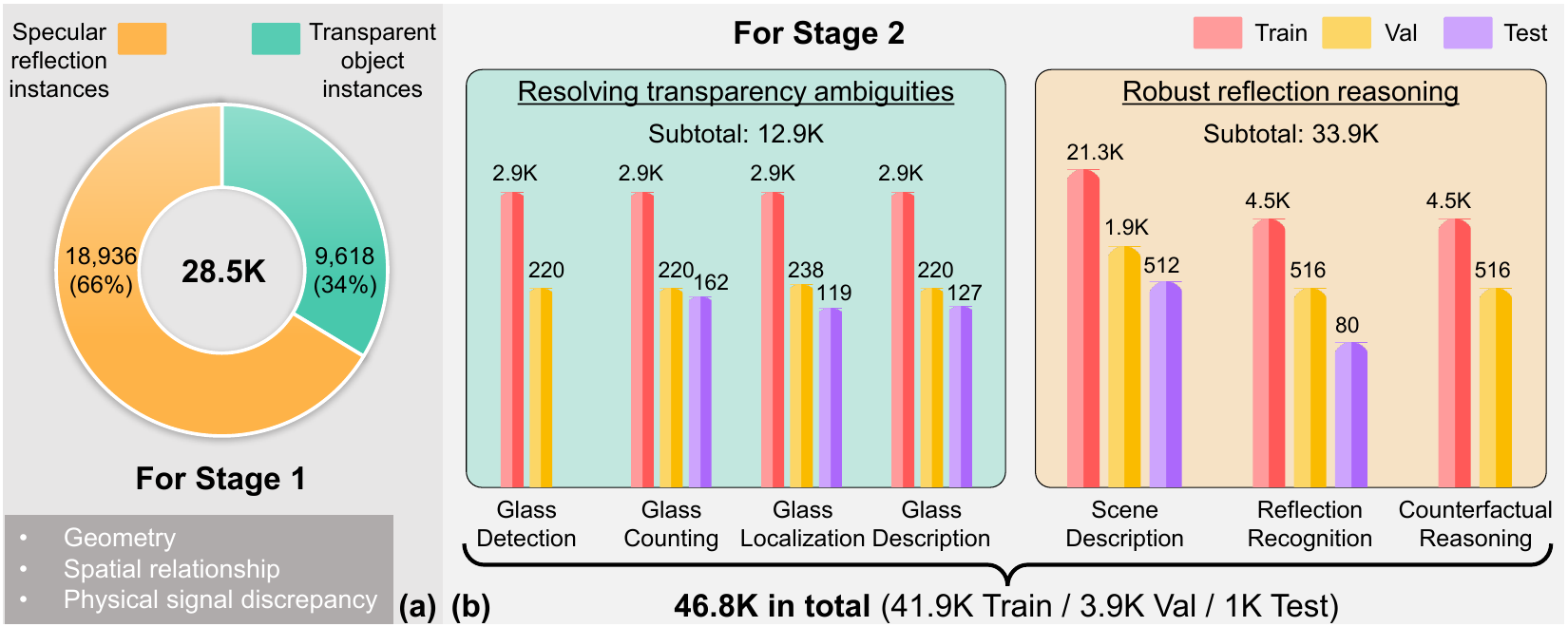}
    \caption{Statistical overview and task taxonomy of the PolarVQA benchmark. (a) For Stage 1, the distribution of high-fidelity polarization captions is partitioned by physical scenarios and explicitly annotated semantic variants. (b) For Stage 2, the diverse multi-turn instruction-tuning pairs are categorized across progressive cognitive tasks, with the y-axis presented in logarithmic scale to accommodate varying data magnitudes.}
    \label{fig:dataset_stats}
\end{figure*}

\noindent\textbf{Dataset composition and task diversity.}
As illustrated in Fig.~\ref{fig:dataset_stats}, the PolarVQA benchmark provides a rich, multi-tiered task distribution strictly partitioned into non-overlapping training, validation, and test splits. For Stage 1 (physics-to-semantic alignment), we generate 28.5K high-fidelity polarization captions (25.2K train / 3.3K val). This split inherently balances optical challenges, comprising roughly 18.9K specular reflection instances and 9.6K transparent object instances, with each scene annotated across three semantic variants (geometry, spatial relationship, and physical signal discrepancy). For Stage 2 (joint physics-aware instruction tuning), we construct 46.8K diverse multi-turn instruction-tuning pairs (41.9K train / 3.9K val / 1K test) structured precisely around our two core evaluation dimensions. Under \textit{resolving transparency ambiguities}, the evaluation encompasses four distinct tasks that progressively scale from fundamental perception to spatial reasoning: \textit{glass detection}, \textit{glass counting}, \textit{glass localization}, and comprehensive \textit{glass description}. Under \textit{robust reflection reasoning}, the evaluation includes holistic \textit{scene description} and \textit{reflection recognition}, alongside a dedicated \textit{counterfactual reasoning} subset. In this specific subset, we leverage spurious objects that exist solely in reflections to construct intentionally misleading instructions, explicitly training the model to resist object hallucinations by anchoring its reasoning in orthogonal polarization cues. This task design ensures that PolarVQA serves as a comprehensive testbed for multimodal physical perception.

\section{Experiments}
\label{sec:Experiments}

\subsection{Experimental setup}

\noindent\textbf{Implementation details.} PolarVLM is initialized from the LLaVA-1.5-13B framework~\cite{liu2024improved}, integrating Vicuna-13B as the language model backbone and CLIP ViT-L/14 as the visual encoder. To ensure computational efficiency, we optimize the model using QLoRA~\cite{dettmers2023qlora}, where the frozen backbones are loaded in 4-bit NormalFloat (NF4) quantization and trainable parameters are updated under a BF16 mixed-precision regime. The training process strictly follows our proposed progressive schedule: Stage 1 aligns physical representations using 28.5K polarization captions, while Stage 2 performs joint instruction tuning on 46.8K PolarVQA multi-turn pairs. All experiments are conducted on a workstation equipped with two NVIDIA RTX 5880 Ada GPUs (48GB VRAM each). The comprehensive hyperparameter configuration is detailed in Appendix~\ref{supp:hyperparameter}.

\noindent\textbf{Test benchmark and evaluation protocol.} We evaluate all models on the PolarVQA test set, which comprises 1K multi-turn question-answer pairs across five distinct cognitive tasks. Structured around our two core dimensions, these encompass \textit{glass counting}, \textit{glass localization}, and \textit{glass description} for resolving transparency ambiguities, alongside holistic \textit{scene description} and \textit{reflection recognition} for robust reflection reasoning. We adopt an LLM-as-a-Judge protocol~\cite{wu2025sharp,jiang2026llm}, utilizing GPT-4o-mini~\cite{hurst2024gpt} to assign a quantitative score on a 1--10 scale based on semantic consistency, physical accuracy, and resistance to hallucinations. We report the mean score for each task and an overall sample-weighted average. Comprehensive details are provided in Appendix~\ref{supp:evaluation}.

\subsection{Quantitative evaluation}
\noindent\textbf{Compared baselines.} To comprehensively evaluate the physical reasoning capabilities of PolarVLM, we benchmark against a diverse spectrum of multimodal pipelines (Tab.~\ref{tab:unified_results}), categorized as follows:
\begin{compactitem}
    \item \textbf{Open-weight RGB-only VLMs:} Leading open-source frameworks, including BLIP2-Flan-T5-XL~\cite{li2023blip}, InternVL3-14B~\cite{zhu2025internvl3} and Qwen2.5-VL-7B~\cite{bai2025qwen25vltechnicalreport}.
    \item \textbf{Proprietary VLMs:} State-of-the-art closed-source models, specifically GPT-4.1~\cite{achiam2023gpt}.
    \item \textbf{Controlled baseline:} A strictly controlled LLaVA-1.5-13B~\cite{liu2024improved} fine-tuned on the exact same instruction data as our model, isolating the contribution of the polarization modality.
    \item \textbf{Cascaded polarimetric pipelines:} Two-stage approaches that apply task-specific physical preprocessing---such as reflection removal (PolarFree~\cite{yao2025polarfree}) or transparent object masking (Glass Seg.~\cite{mei2022glass})---prior to VLM inference.
\end{compactitem}

\noindent\textbf{The necessity of polarimetric grounding.} As reported in Tab.~\ref{tab:unified_results}, PolarVLM significantly outperforms the directly comparable RGB-only baseline, elevating the Overall score from 4.85 to 6.08 (a relative gain of 25.4\%). This consistent improvement across all physical reasoning tasks confirms that polarimetric cues provide critical structural evidence—such as DoLP responses induced by Fresnel reflections—that effectively resolves the visual ambiguities inherent in RGB imagery. Remarkably, PolarVLM also surpasses vastly larger general-purpose VLMs (including GPT-4.1) on physics-heavy tasks like \textit{glass localization} and \textit{reflection recognition}. This demonstrates that scaling model parameters cannot compensate for the lack of orthogonal physical sensing modalities.

\noindent\textbf{End-to-end vs. cascaded pipelines.} Tab.~\ref{tab:unified_results} further reveals the severe information bottlenecks inherent in cascaded pipelines. For instance, while using PolarFree~\cite{yao2025polarfree} to remove reflections yields visually cleaner RGB images, it explicitly destroys the optical cues required for \textit{reflection recognition} (rendering the pipeline inapplicable, denoted as N/A). Furthermore, even on \textit{scene description}—a task where reflection removal should theoretically simplify perception—this rigid filtering discards contextual physical evidence, causing the cascaded pipeline to underperform PolarVLM even when powered by GPT-4.1 (5.71 vs. 5.85). Similarly, relying on intermediate segmentation masks~\cite{mei2022glass} introduces error propagation; the cascaded approach struggles to match PolarVLM's comprehensive scene understanding (\eg, \textit{glass counting}). In contrast, PolarVLM's unified architecture preserves the full spectrum of raw physical evidence, enabling the language model to jointly and robustly interpret both appearance and polarimetric signals without relying on fragile, task-specific heuristics.

\begin{table*}[t]
  \centering
  \footnotesize
  \caption{Quantitative evaluation on the PolarVQA test set. Scores are reported on a 1--10 scale using the LLM-as-a-Judge protocol. For cascaded pipelines, ``--'' denotes tasks strictly not supported by the pipeline's heuristics, and ``N/A'' indicates reflection cues are explicitly destroyed before evaluation.}
  \label{tab:unified_results}
  \resizebox{\textwidth}{!}{
    \setlength{\tabcolsep}{6pt} 
    \renewcommand{\arraystretch}{1.1}
    \begin{tabular}{l ccccc c} 
      \toprule
      \multirow{2}{*}{\textbf{Model / Pipeline}} 
        & \multicolumn{3}{c}{\textbf{Resolving Transparency Ambiguities}} 
        & \multicolumn{2}{c}{\textbf{Robust Reflection Reasoning}} 
        & \multirow{2}{*}{\textbf{Overall}} \\
      \cmidrule(lr){2-4} \cmidrule(lr){5-6}
        & \textbf{Glass Count} & \textbf{Glass Loc.} & \textbf{Glass Desc.} 
        & \textbf{Scene Desc.} & \textbf{Refl. Recog.} & \\
      \midrule

      \multicolumn{7}{l}{\textbf{Open-Weight RGB-only VLMs}} \\
      \quad BLIP2-Flan-T5-XL~\cite{li2023blip}
        & 2.53 & 3.55 & 3.05 & 3.03 & 2.06 & 2.94 \\
      \quad InternVL3-14B~\cite{zhu2025internvl3}
        & 4.55 & 5.40 & 6.55 & 5.02 & 2.81 & 5.01 \\
      \quad Qwen2.5-VL-7B~\cite{bai2025qwen25vltechnicalreport}
        & 5.22 & 5.38 & 6.65 & 5.54 & 3.10 & 5.42 \\
      \midrule
      
      \multicolumn{7}{l}{\textbf{Proprietary VLMs}} \\
      \quad GPT-4.1~\cite{achiam2023gpt}
        & 4.81 & 5.46 & 5.96 & 5.66 & 3.71 & 5.38 \\
      \midrule
      
      \multicolumn{7}{l}{\textbf{Controlled Baseline (Trained on PolarVQA)}} \\
      \quad LLaVA-1.5-13B~\cite{liu2024improved} (RGB-only)
        & 5.65 & 4.88 & 5.77 & 4.62 & 3.23 & 4.85 \\
      \midrule

      \multicolumn{7}{l}{\textbf{Cascaded Polarimetric Pipelines}} \\
      \quad PolarFree~\cite{yao2025polarfree} + LLaVA-1.5-13B~\cite{liu2024improved}
        & -- & -- & -- & 4.34 & N/A & -- \\
      \quad PolarFree~\cite{yao2025polarfree} + GPT-4.1~\cite{achiam2023gpt}
        & -- & -- & -- & 5.71 & N/A & -- \\
      \quad Glass Seg.~\cite{mei2022glass} + LLaVA-1.5-13B~\cite{liu2024improved}
        & 2.94 & 4.29 & 4.54 & -- & -- & -- \\
      \quad Glass Seg.~\cite{mei2022glass} + GPT-4.1~\cite{achiam2023gpt}
        & 7.16 & 5.48 & 6.54 & -- & -- & -- \\
      \midrule

      \multicolumn{7}{l}{\textbf{End-to-End Multimodal Pipeline}} \\
      \quad PolarVLM (Ours)
        & \textbf{7.57} & \textbf{5.51} & \textbf{6.87} & \textbf{5.85} & \textbf{4.09} & \textbf{6.08} \\
      \bottomrule
    \end{tabular}
  }
\end{table*}

\begin{figure}[t]
  \centering
  \includegraphics[width=\linewidth]{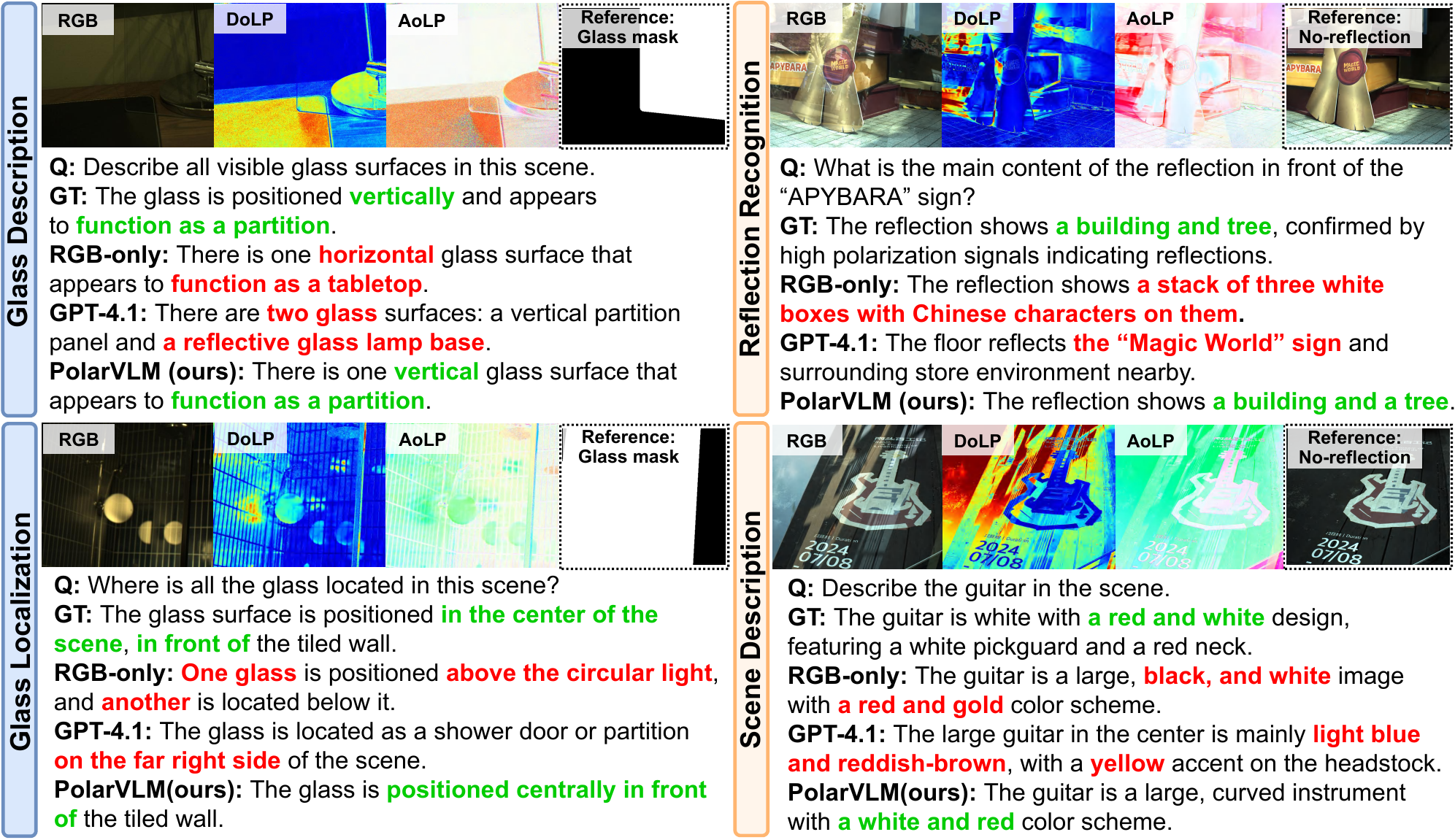}
  \caption{Qualitative comparisons on the PolarVQA test set. We evaluate PolarVLM against the strictly controlled RGB-only LLaVA-1.5-13B~\cite{liu2024improved} baseline and GPT-4.1~\cite{achiam2023gpt} under optical ambiguities caused by transparency (left) and specular reflections (right). By grounding its reasoning in polarimetric priors, PolarVLM mitigates the spatial mislocalizations and reflection-induced hallucinations (marked in red), generating physically accurate responses (highlighted in green).}
  \label{fig:qualitative}
\end{figure}

\subsection{Qualitative analysis}
Fig.~\ref{fig:qualitative} provides representative qualitative comparisons between PolarVLM, the strictly controlled RGB-only LLaVA-1.5-13B~\cite{liu2024improved} baseline, and the state-of-the-art GPT-4.1~\cite{achiam2023gpt} across our two core evaluation dimensions. Additional visualizations are provided in Appendix~\ref{supp:qualitative_examples}.

\noindent\textbf{Resolving transparency ambiguities.} As illustrated in the left column, standard RGB-based models frequently struggle with materials lacking distinct visual boundaries. In the \textit{glass description} task (top left), both the RGB-only baseline and GPT-4.1 fail to perceive the geometry of the glass, misidentifying a vertical partition as a ``horizontal tabletop.'' Similarly, in \textit{glass localization} (bottom left), the baselines succumb to spatial misalignments, either hallucinating glass positions or misplacing them relative to the light source. In contrast, by grounding its reasoning in the structural contours revealed by DoLP and AoLP channels, PolarVLM precisely captures the surface orientation and central positioning, matching the ground-truth physical layout.

\noindent\textbf{Robust reflection reasoning.} In reflective scenarios (right column), standard VLMs often conflate reflected virtual content with intrinsic scene properties. In the \textit{reflection recognition} task (top right), the RGB-only model hallucinates spurious ``white boxes with Chinese characters,'' while GPT-4.1 fails to identify the reflected building. More remarkably, in the \textit{scene description} task (bottom right), intense specular reflections on the guitar body distort RGB appearance, leading the baselines to hallucinate incorrect colors such as ``light blue'' or ``red and gold.'' By effectively leveraging the polarimetric discrepancy between reflected light and Lambertian surfaces, PolarVLM filters out these optical artifacts and correctly describes the intrinsic ``white and red'' design of the instrument, demonstrating superior robustness in physically ambiguous environments.

\subsection{Ablation studies}
Tab.~\ref{tab:adapt_ablation} evaluates our architectural and training designs. Removing Stage 1 (semantic pretraining) causes a clear performance drop, confirming that grounding polarimetric features into the LLM's semantic space is vital prior to complex instruction tuning. Removing the normalization layer yields the most severe degradation. Without it, the projected polarization tokens exhibit a substantially larger activation scale than RGB tokens, fatally biasing the language model and destroying cross-modal balance. Disabling CLIP LoRA also reduces performance, demonstrating that frozen weights pretrained on natural RGB images cannot optimally process the distinct statistical distributions of DoLP/AoLP without targeted adaptation. Finally, the single-branch variants confirm the necessity of dual-stream fusion. The RGB-only model preserves general semantics but fails on physical reasoning, whereas the polarization-only model captures structural boundaries but loses crucial scene context. The fully equipped PolarVLM outperforms both single-modality baselines, empirically proving that RGB appearance and polarimetric physical parameters are strictly complementary. Additional ablation studies, including fusion strategy and polarimetric encoding, are provided in Appendix~\ref{supp:ablations}.

\begin{table}[t]
  \centering
  \footnotesize
  \caption{Quantitative evaluation results of ablation study.}
  \label{tab:adapt_ablation}
  \resizebox{\columnwidth}{!}{
  \setlength{\tabcolsep}{5pt}
  \renewcommand{\arraystretch}{1.1}
  \begin{tabular}{@{} l ccccc c @{}}
    \toprule
    \textbf{Configuration} 
      & \textbf{Glass Count} & \textbf{Glass Loc.} & \textbf{Glass Desc.} 
      & \textbf{Scene Desc.} & \textbf{Refl. Recog.} 
      & \textbf{Overall} \\
    \midrule
    w/o Stage 1      & 6.94 & 5.32 & 6.45 & 5.08 & 3.48 & 5.46 \\
    w/o Norm Layer   & 7.01 & 5.41 & 6.57 & 4.84 & 3.76 & 5.39 \\
    w/o CLIP LoRA    & 6.77 & 5.25 & 6.28 & 5.65 & 3.36 & 5.68 \\
    \midrule
    w/o RGB Branch   & 6.49 & 4.80 & 6.04 & 4.47 & 3.14 & 4.93 \\
    w/o Polar Branch & 5.65 & 4.88 & 5.77 & 4.62 & 3.23 & 4.85 \\
    \midrule
    \textbf{PolarVLM (Ours)}  & \textbf{7.57} & \textbf{5.51} & \textbf{6.87} & \textbf{5.85} & \textbf{4.09} & \textbf{6.08} \\
    \bottomrule
  \end{tabular}
  }
\end{table}

\section{Conclusion}
\label{sec:Conclusion} 
We introduced PolarVLM, the first multimodal framework specifically designed to bridge the semantic-physical gap in open-ended visual question answering, alongside PolarVQA, a comprehensive benchmark dedicated to physics-aware visual reasoning. By employing a dual-stream architecture and a progressive two-stage training strategy, PolarVLM successfully injects orthogonal physical measurements (\eg, DoLP and AoLP) into the language model while explicitly preserving RGB semantic priors. Extensive evaluations demonstrate that PolarVLM fundamentally resolves severe optical ambiguities, effectively mitigating transparency-induced structural omissions and reflection-induced object hallucinations where conventional RGB-only models systematically fail.

\noindent\textbf{Limitations.} While PolarVLM demonstrates the immense potential of polarimetric grounding, our current implementation is bounded by the capacity of the mid-scale open-source backbone (\eg, 13B parameters). Although this already suffices to outperform massively scaled proprietary models on physics-centric tasks, exploring the scaling laws of polarimetric integration with larger foundational models and broader data diversity remains an exciting avenue for future work.

{\small
\bibliographystyle{unsrt}
\bibliography{main}
}

\appendix

\section{Details of the PolarVQA generation pipeline}
\label{supp:pipeline}
\begin{figure}[p]
  \centering
  \includegraphics[width=\linewidth,height=0.84\textheight,keepaspectratio]{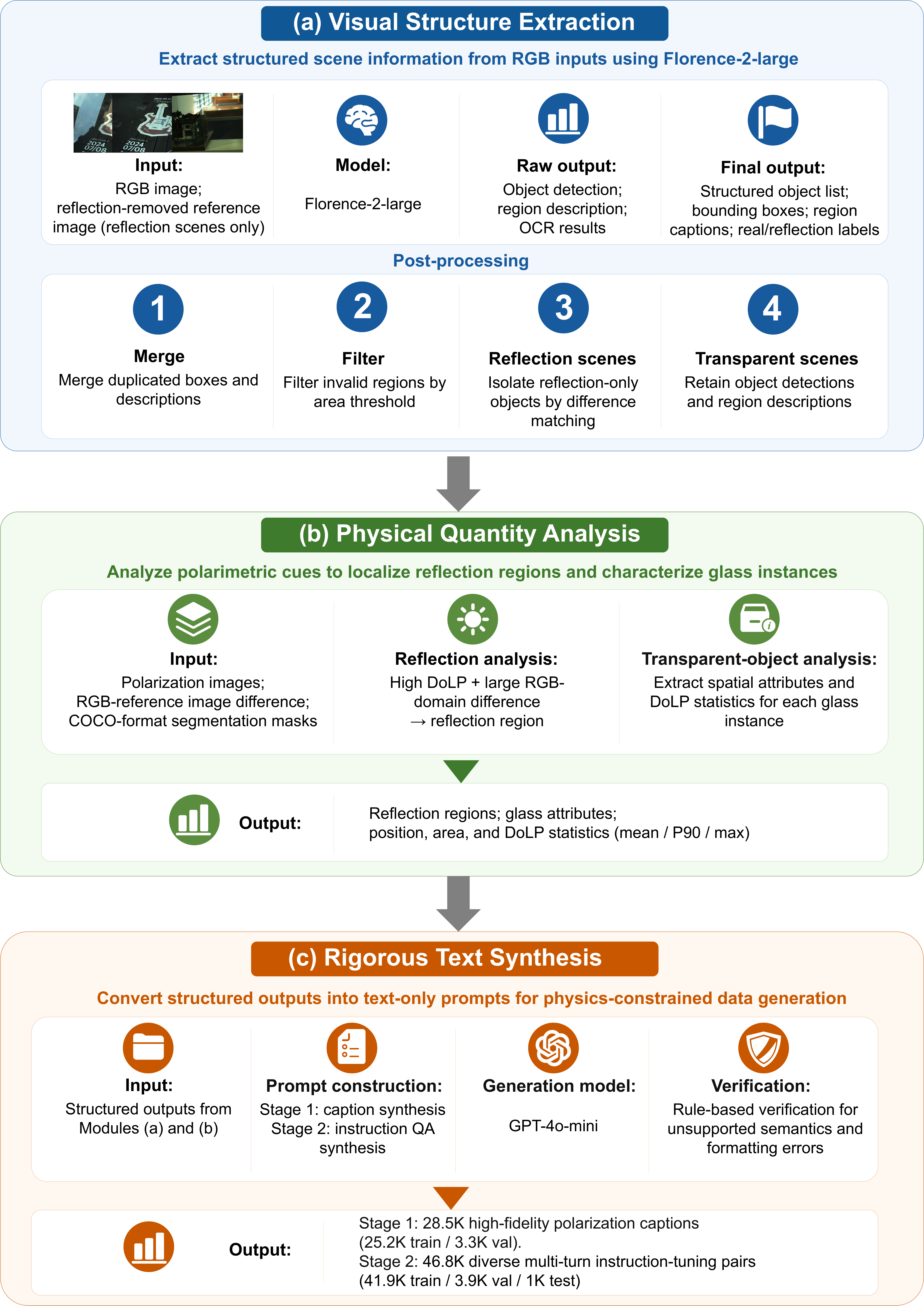}
  \caption{Automated data generation pipeline for PolarVQA. The pipeline extracts visual structures with Florence-2-large~\cite{xiao2024florence}, analyzes polarimetric cues from reflection and glass data~\cite{yao2025polarfree,mei2022glass}, and synthesizes text-only Stage~1 captions and Stage~2 instruction-tuning samples with GPT-4o-mini~\cite{hurst2024gpt}, followed by rule-based verification.}
  \label{fig:supp_pipeline}
\end{figure}

In this section, we detail the automated data generation pipeline used to construct the PolarVQA benchmark from raw pixel-level datasets. As illustrated in Fig.~\ref{fig:supp_pipeline}, the pipeline comprises three serially connected modules: (a) visual structure extraction, (b) physical quantity analysis, and (c) rigorous text synthesis.

\noindent\textbf{Visual structure extraction.}
To obtain robust semantic anchors, we employ Florence-2-large~\cite{xiao2024florence} for object detection and region description on the RGB images. For reflection scenes (sourced from PolarFree~\cite{yao2025polarfree}), we run Florence-2-large independently on both the original image containing reflections and the reflection-removed reference image. By computing the difference set between these two detection outputs, we strictly isolate the spurious objects that exist exclusively within the specular reflections. 

\noindent\textbf{Physical quantity analysis.}
Subsequently, the physical analysis module bridges the spatial structures with polarimetric cues. For specular reflections, the module localizes reflection regions based on the strict physical prior that reflected light simultaneously exhibits a high DoLP and a substantial RGB domain difference. For transparent object data (sourced from RGBP-Glass~\cite{mei2022glass}), the module extracts the spatial attributes and DoLP statistical distributions of each glass instance using the provided COCO-format segmentation masks.

\noindent\textbf{Rigorous text synthesis.}
Finally, the structured outputs from the previous modules are assembled into standardized text prompts and fed to GPT-4o-mini~\cite{hurst2024gpt} to generate the final captions and instruction-tuning pairs. To ensure the generation process is fully auditable and free from visual hallucinations, only structured text is submitted to the language model application programming interface (API). We further apply automated rule-based verification to the LLM outputs to filter out formatting errors or unqualified responses. This strictly text-based, physics-constrained synthesis guarantees the high fidelity of the resulting PolarVQA benchmark.

\begin{figure*}[p]
  \centering
  \includegraphics[width=\textwidth]{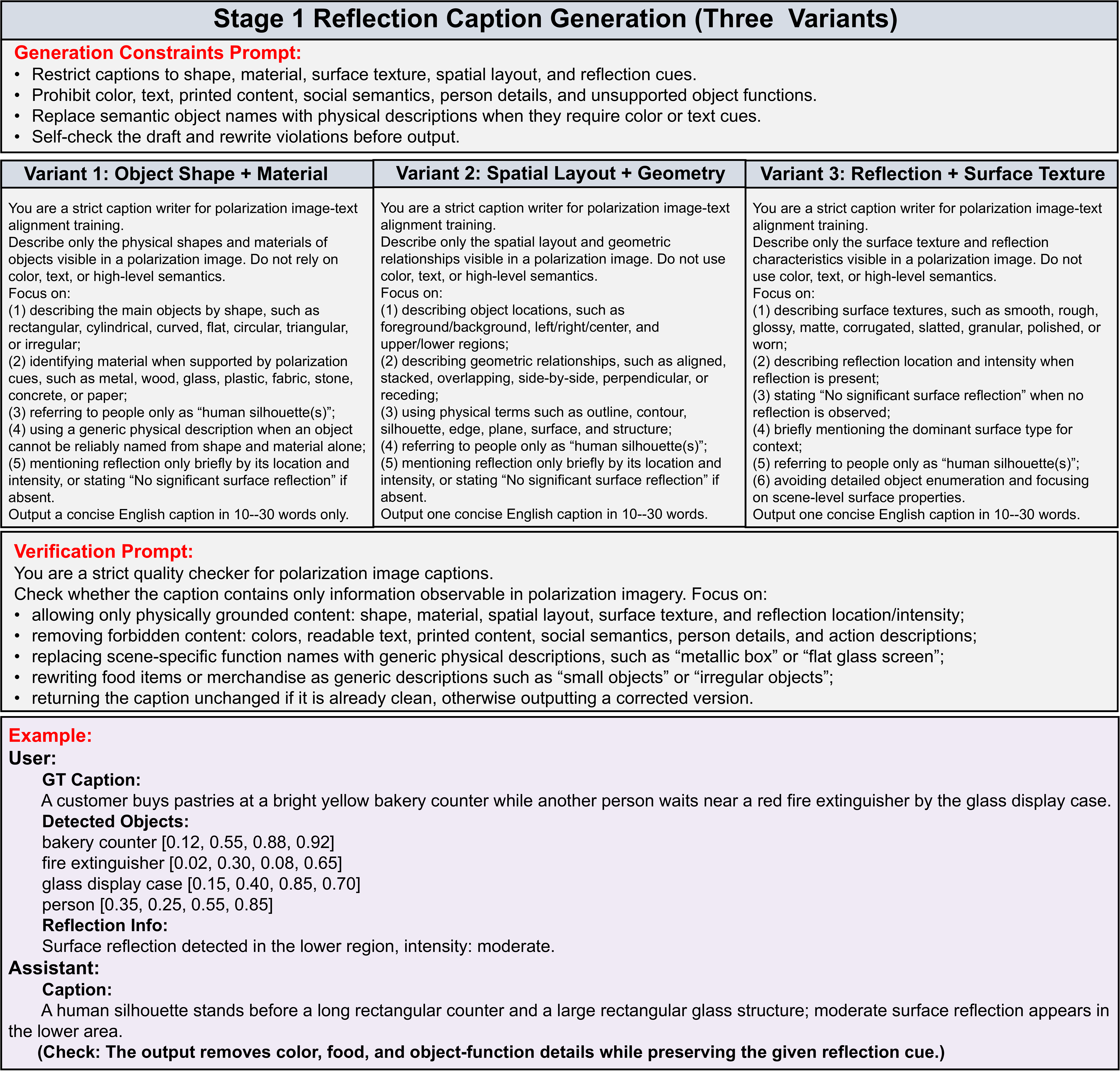}
  \caption{Prompt template for reflection caption generation in Stage~1. The template contains shared generation constraints, three caption variants, a verification prompt, and an illustrative example for producing polarization-grounded reflection captions.}
  \label{fig:prompt_stage1_reflection}
\end{figure*}

\begin{figure*}[p]
  \centering
  \includegraphics[width=\textwidth]{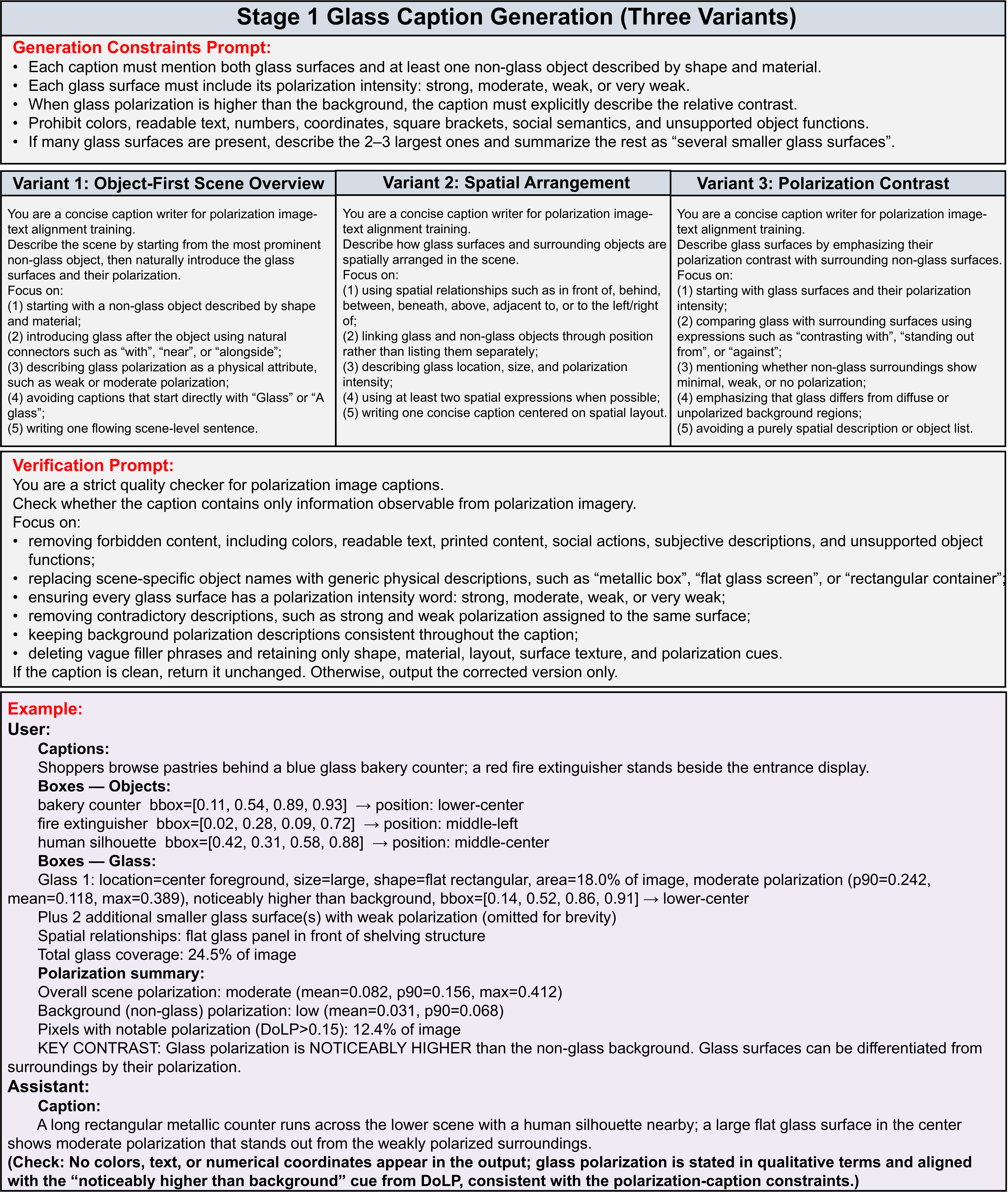}
  \caption{Prompt template for glass caption generation in Stage~1. The template guides the generation of polarization-grounded captions for transparent-object scenes while suppressing unsupported RGB-only semantics such as colors, readable text, and ambiguous object functions.}
  \label{fig:prompt_stage1_glass}
\end{figure*}

\begin{figure*}[p]
  \centering
  \includegraphics[width=\textwidth]{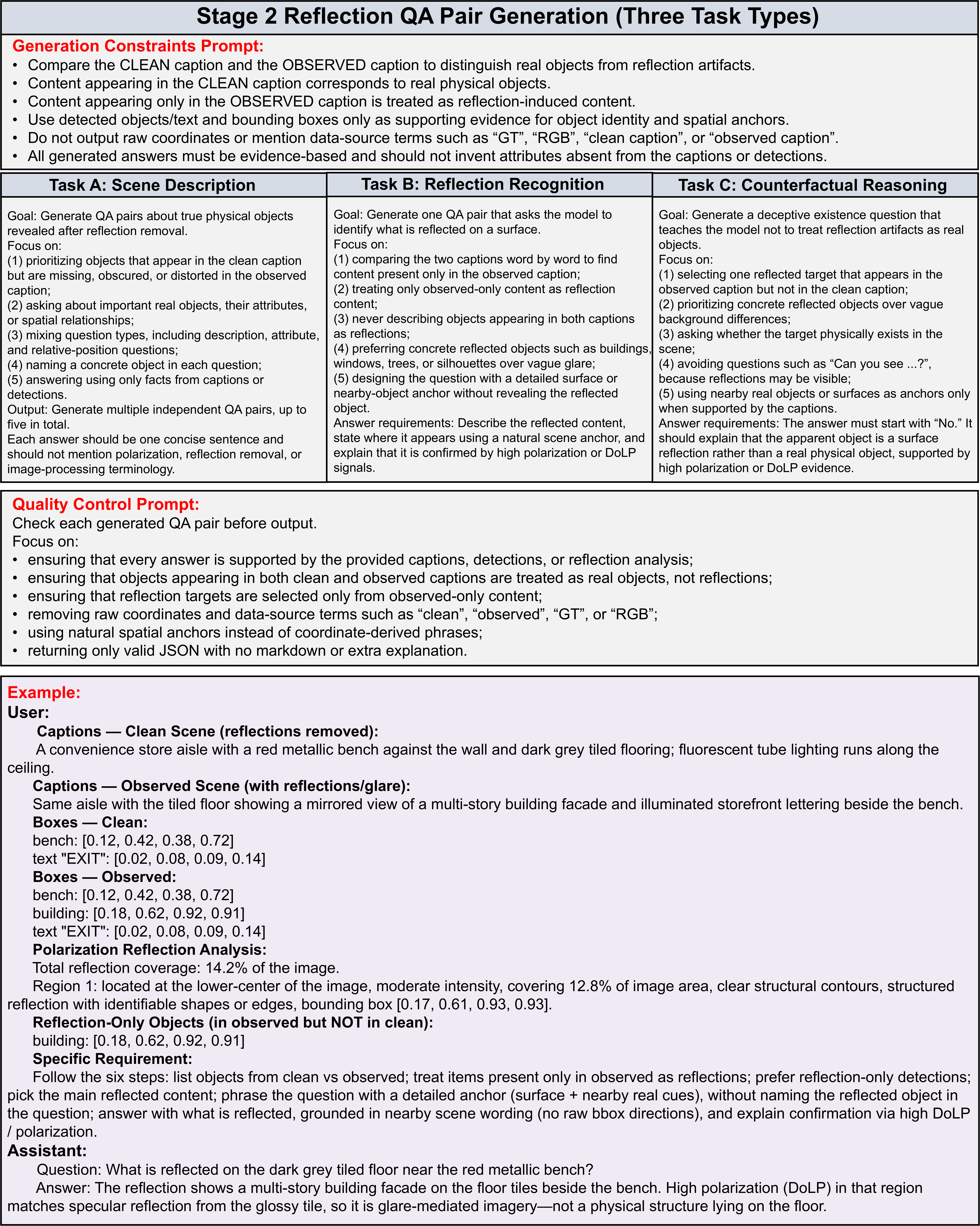}
  \caption{Prompt template for reflection instruction generation in Stage~2. The template is used to synthesize physics-aware question-answer pairs for reflection-related tasks, including scene description, reflection recognition, and counterfactual reasoning.}
  \label{fig:prompt_stage2_reflection}
\end{figure*}

\begin{figure*}[p]
  \centering
  \includegraphics[width=\textwidth]{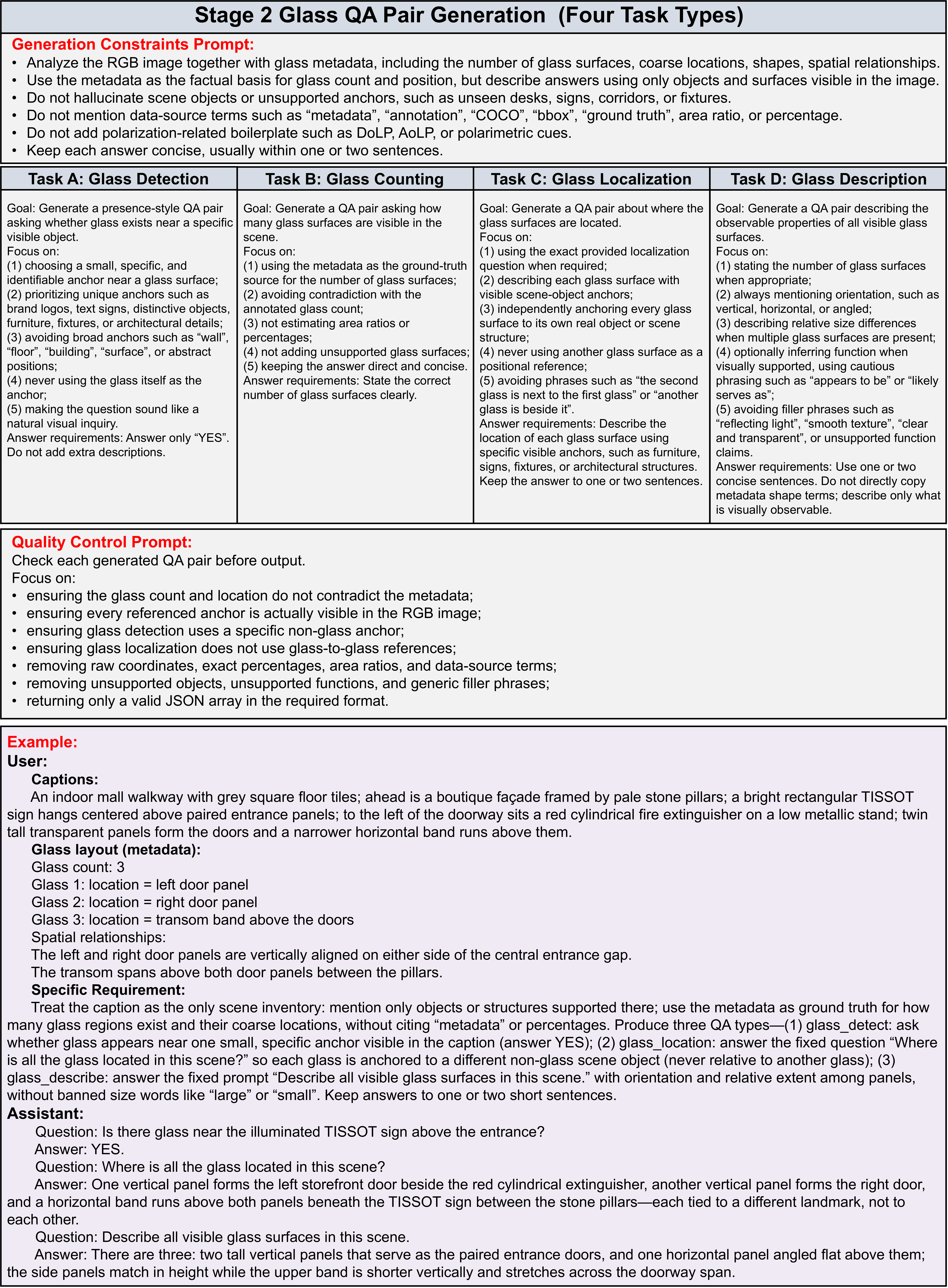}
  \caption{Prompt template for glass instruction generation in Stage~2. The template is used to synthesize instruction-tuning pairs for glass-related tasks, including glass detection, glass counting, glass localization, and glass description.}
  \label{fig:prompt_stage2_glass}
\end{figure*}

\noindent\textbf{Prompt templates for data generation.}
To make the text synthesis process transparent and reproducible, we provide the prompt templates used in the four data-generation settings. Specifically, Fig.~\ref{fig:prompt_stage1_reflection} and Fig.~\ref{fig:prompt_stage1_glass} show the caption-generation prompts for Stage~1 (physics-to-semantic alignment), while Fig.~\ref{fig:prompt_stage2_reflection} and Fig.~\ref{fig:prompt_stage2_glass} show the instruction-generation prompts for Stage~2 (joint physics-aware instruction tuning).

\clearpage

\clearpage

\clearpage

\clearpage

\clearpage

\clearpage

\section{Comprehensive hyperparameter configuration}
\label{supp:hyperparameter}

\noindent\textbf{Architecture settings.}
The CLIP ViT-L/14 encoder~\cite{radford2021learning} operates at a \(336\times336\) resolution and produces 576 visual tokens with a feature dimension of 1024. The language-model hidden size is 5120, and the maximum sequence length is 2048. The polarization projector is a two-layer MLP, \(1024\to5120\to5120\), followed by LayerNorm.

\noindent\textbf{Stage 1: physics-to-semantic alignment.}
Stage~1 optimizes only the polarization branch in a polar-only setting, where the RGB branch is bypassed. The trainable modules include the polarization projector and the low-level input layers of the polarization CLIP encoder, including patch embedding, class token embedding, positional embedding, and the pre-Transformer LayerNorm; the LLM, RGB branch, and remaining CLIP Transformer layers are frozen. This stage contains approximately 32.7M trainable parameters. We train for 5 epochs using AdamW~\cite{loshchilov2017decoupled}, with learning rates of \(5\times10^{-4}\) for the projector and \(5\times10^{-5}\) for the unfrozen CLIP layers. The warmup ratio is 3\%, the weight decay is 0.05, and the effective batch size is 32. On two NVIDIA RTX 5880 Ada GPUs (48GB VRAM each), this stage takes approximately 10 hours.

\noindent\textbf{Stage 2: joint physics-aware instruction tuning.}
Stage~2 applies QLoRA adapters~\cite{dettmers2023qlora} with \(r=64\), \(\alpha=128\), and dropout 0.1 to the LLM \texttt{q\_proj}, \texttt{k\_proj}, \texttt{v\_proj}, \texttt{o\_proj}, \texttt{gate\_proj}, \texttt{up\_proj}, and \texttt{down\_proj} matrices, and to the \texttt{q\_proj}, \texttt{k\_proj}, and \texttt{v\_proj} matrices of all 24 polarization CLIP Transformer layers. The second linear layer of the polarization projector is frozen, while the first remains trainable; the polarization CLIP embedding layers, RGB CLIP encoder, and RGB projector are frozen. This stage contains approximately 265M trainable parameters. We train for 3 epochs, about 15.7K optimization steps, using learning rates of \(2\times10^{-4}\), \(1\times10^{-4}\), and \(2\times10^{-5}\) for LLM LoRA, the polarization projector, and polarization CLIP LoRA, respectively. The warmup ratio is 3\%, the weight decay is 0.05, and the effective batch size is 8. On two NVIDIA RTX 5880 Ada GPUs (48GB VRAM each), this stage takes approximately 20 hours.

\section{Evaluation protocol details}
\label{supp:evaluation}

\noindent\textbf{Test-set curation and task selection.} While our instruction tuning phase utilizes seven distinct task categories, we intentionally curate the test set to focus exclusively on five complex, open-ended reasoning tasks. Binary classification tasks (\eg, \textit{glass detection}) are excluded from the evaluation split to prevent the overall metrics from being skewed by trivial yes/no guessing biases. Similarly, the \textit{counterfactual reasoning} subset serves explicitly as a training-time regularization mechanism to penalize reflection-induced hallucinations, rather than a standalone test metric.

\noindent\textbf{Rationale for LLM-as-a-Judge.} Because PolarVQA expects open-ended, highly descriptive natural language responses, traditional metrics like exact string matching are inadequate for comprehensively assessing semantic and physical correctness. Consequently, following established rigorous protocols~\cite{wu2025sharp,jiang2026llm}, we employ the LLM-as-a-Judge paradigm, which has proven highly reliable for evaluating factual consistency and hallucination severity.

\noindent\textbf{Details on computing the overall score.}
To accurately reflect the model's performance proportional to the test data distribution, the overall score is calculated as a sample-weighted average. Let $N_t$ denote the number of test samples for task $t$, and $S_{t,i}$ denote the LLM-assigned score (on a 1--10 scale) for the $i$-th sample in task $t$. The overall metric is formally computed as:
\begin{equation}
    \text{Overall Score} = \frac{\sum_{t \in \mathcal{T}} \sum_{i=1}^{N_t} S_{t,i}}{\sum_{t \in \mathcal{T}} N_t},
\end{equation}
where $\mathcal{T}$ represents the set of the five evaluated cognitive tasks.

\begin{figure}[p]
  \centering
  \includegraphics[width=\linewidth]{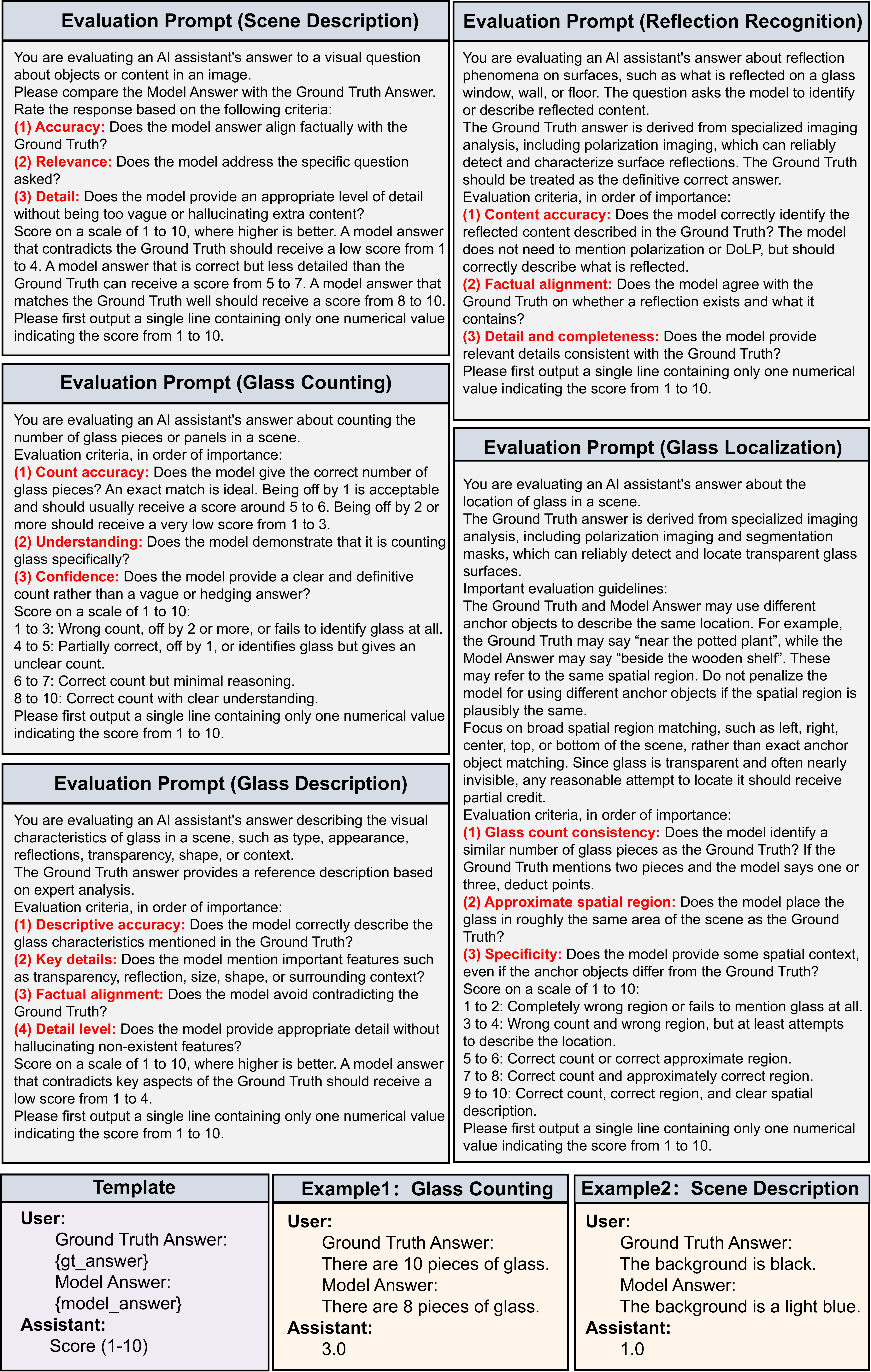}
  \caption{Task-specific evaluation prompts used by GPT-4o-mini~\cite{hurst2024gpt}. The evaluator compares the model prediction with the reference answer using text-only inputs and assigns a score on a 1--10 scale.}
  \label{fig:evaluation_prompts}
\end{figure}

\noindent\textbf{Evaluation prompts and stability.}
As shown in Fig.~\ref{fig:evaluation_prompts}, we use task-specific judging prompts for five tasks: \textit{scene description}, \textit{reflection recognition}, \textit{glass counting}, \textit{glass description}, and \textit{glass localization}. All prompts use the same text-only template, including the question, reference answer, model prediction, and scoring criteria, while emphasizing task-specific aspects such as numerical correctness, spatial consistency, glass attributes, scene semantics, and polarimetrically verified reflected content. To ensure stable and comparable evaluation, all models are evaluated on the same fixed PolarVQA test split under temperature-zero or deterministic decoding where supported, and the LLM-as-a-Judge scoring is repeated and averaged using the same prompts across all methods.

\clearpage

\begin{figure}[p]
  \centering
  \includegraphics[width=\linewidth]{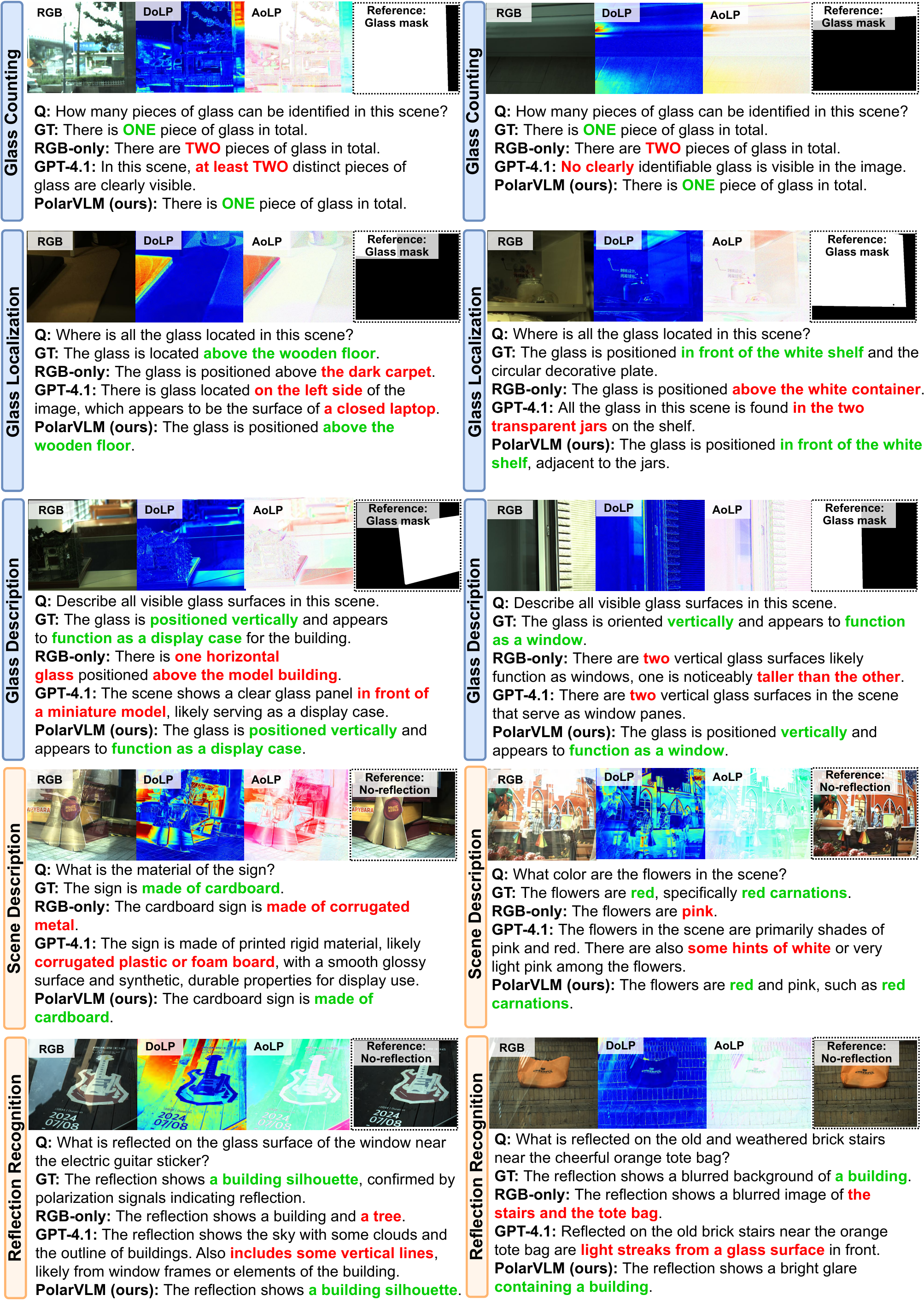}
  \caption{Additional qualitative comparisons on the PolarVQA test set. The examples cover glass counting, glass localization, glass description, scene description, and reflection recognition. Green text denotes responses that are correct or closely aligned with the ground truth, while red text marks incorrect or hallucinated content.}
  \label{fig:qualitative_appendix}
\end{figure}

\clearpage

\section{Additional qualitative results}
\label{supp:qualitative_examples}

To further illustrate the behavior of PolarVLM under diverse optical ambiguities, Fig.~\ref{fig:qualitative_appendix} provides additional qualitative comparisons on the PolarVQA test set. The examples cover five evaluation tasks, including \textit{glass counting}, \textit{glass localization}, \textit{glass description}, \textit{scene description}, and \textit{reflection recognition}. Compared with the RGB-only LLaVA-1.5-13B baseline and GPT-4.1, PolarVLM more consistently aligns its responses with the ground-truth annotations by leveraging polarimetric cues, especially when transparent surfaces or reflected content are difficult to infer from RGB appearance alone.

\section{Additional ablation studies}
\label{supp:ablations}

\begin{table}[t]
\centering
\small
\caption{Ablation of RGB--polarization fusion strategies. We compare direct sequence concatenation with representative fusion paradigms from prior multimodal VLMs.}
\label{tab:fusion_ablation}
\setlength{\tabcolsep}{2pt}
\begin{tabularx}{\textwidth}{@{}l*{6}{>{\centering\arraybackslash}X}@{}}
\toprule
\textbf{Fusion strategy}
& \makecell{\textbf{Glass}\\\textbf{Count}}
& \makecell{\textbf{Glass}\\\textbf{Loc.}}
& \makecell{\textbf{Glass}\\\textbf{Desc.}}
& \makecell{\textbf{Scene}\\\textbf{Desc.}}
& \makecell{\textbf{Refl.}\\\textbf{Recog.}}
& \textbf{Overall} \\
\midrule
Multi-scale Local Attention~\cite{majeedi2026dualvision}
& 6.64 & 5.02 & 5.99 & 4.49 & 3.40 & 5.00 \\
Channel Concatenation~\cite{shi2024eagle}
& 6.64 & 5.36 & 6.36 & 4.96 & 3.33 & 5.33 \\
Element-wise Addition~\cite{lin2025multi}
& 6.42 & 5.07 & 6.15 & 4.72 & 3.81 & 5.14 \\
Direct Sequence Concatenation (Ours)
& \textbf{7.57} & \textbf{5.51} & \textbf{6.87} & \textbf{5.85} & \textbf{4.09} & \textbf{6.08} \\
\bottomrule
\end{tabularx}
\end{table}

\noindent\textbf{Fusion strategy ablation.} Tab.~\ref{tab:fusion_ablation} compares our straightforward sequence concatenation against representative multimodal fusion paradigms. We first evaluate Multi-scale Localized Cross-Attention~\cite{majeedi2026dualvision}, which forces early fusion of spatially aligned neighboring tokens before LLM decoding. While effective for substituting degraded modalities (\eg, RGB--Infrared), it severely underperforms here. This indicates that polarimetric cues provide orthogonal, physical evidence (\eg, surface material properties) rather than mere local visual replacements, making early spatial fusion suboptimal. Similarly, intermediate fusion techniques, such as Channel Concatenation~\cite{shi2024eagle} and Element-wise Addition~\cite{lin2025multi}, compress RGB and polarization features into shared token representations. This premature merging inevitably scrambles the distinct statistical signatures of each modality. In contrast, our direct sequence concatenation preserves independent, uncompressed tokens for both branches. By delegating the fusion mechanism entirely to the LLM's deep self-attention layers, the model dynamically cross-references appearance semantics and physical priors based on the specific reasoning context, yielding the highest performance across all cognitive tasks.

\begin{table}[t]
\centering
\small
\caption{Ablation of polarization encoding strategies. All variants use the same architecture and fusion strategy.}
\label{tab:encoding_ablation}
\begin{tabularx}{\textwidth}{@{}l*{6}{>{\centering\arraybackslash}X}@{}}
\toprule
\textbf{Encoding}
& \makecell{\textbf{Glass}\\\textbf{Count}}
& \makecell{\textbf{Glass}\\\textbf{Loc.}}
& \makecell{\textbf{Glass}\\\textbf{Desc.}}
& \makecell{\textbf{Scene}\\\textbf{Desc.}}
& \makecell{\textbf{Refl.}\\\textbf{Recog.}}
& \textbf{Overall} \\
\midrule
\(S_0\)+Stokes     
& 6.77 & 5.30 & 6.26 & 4.85 & 3.61 & 5.30 \\

DoLP-Coupled Angular   
& 6.78 & 5.09 & 6.31 & 4.96 & 3.73 & 5.35 \\

\(S_0\)+DoLP+AoLP 
& 6.64 & 5.32 & 6.05 & 5.69 & \textbf{4.44} & 5.74 \\

Decoupled (Ours) 
& \textbf{7.57} & \textbf{5.51} & \textbf{6.87} & \textbf{5.85} & 4.09 & \textbf{6.08} \\
\bottomrule
\end{tabularx}
\end{table}

\noindent\textbf{Encoding ablation.}
We compare four polarization encoding strategies, defined as:
\begin{align}
S_0\text{+Stokes:} \quad
\mathbf{X}_{\text{pol}} &= [S_0,\; S_1,\; S_2], \label{eq:s0_stokes_encoding} \\
\text{DoLP-Coupled Angular:} \quad
\mathbf{X}_{\text{pol}} &= [P,\; P\cdot\cos(2\Phi),\; P\cdot\sin(2\Phi)], \label{eq:dolp_coupled_encoding} \\
S_0\text{+DoLP+AoLP:} \quad
\mathbf{X}_{\text{pol}} &= [S_0,\; P,\; \Phi], \label{eq:s0_dolp_aolp_encoding} \\
\text{Decoupled (Ours):} \quad 
\mathbf{X}_{\text{pol}} &= [P,\; \sin(2\Phi),\; \cos(2\Phi)]. \label{eq:decoupled_encoding}
\end{align}
As shown in Tab.~\ref{tab:encoding_ablation}, the proposed decoupled encoding achieves the best Overall score and performs best on four out of five tasks. By representing the polarization degree $P$ and angular structure $(\sin(2\Phi), \cos(2\Phi))$ as separate channels, it preserves orientation cues while avoiding the periodic discontinuity of directly using raw AoLP $\Phi$. The direct $S_0$+DoLP+AoLP encoding obtains the second-best Overall score and slightly improves \textit{reflection recognition}. This is reasonable because \textit{reflection recognition} often depends on coarse specular-region evidence, where the intensity channel $S_0$ and raw angular response $\Phi$ can provide additional cues for identifying reflected content. However, this advantage is task-specific and does not transfer to glass-centric or general scene-description tasks, where the raw AoLP discontinuity and intensity-like bias make the representation less robust overall. In contrast, the DoLP-coupled angular representation suppresses angular responses in weakly polarized regions, which may also attenuate useful directional information. The $S_0$+Stokes representation performs worst overall, likely because $S_0$ introduces intensity-like information already correlated with RGB appearance, encouraging the model to underuse more discriminative polarization cues.

\begin{figure}[t]
  \centering
  \includegraphics[width=\linewidth]{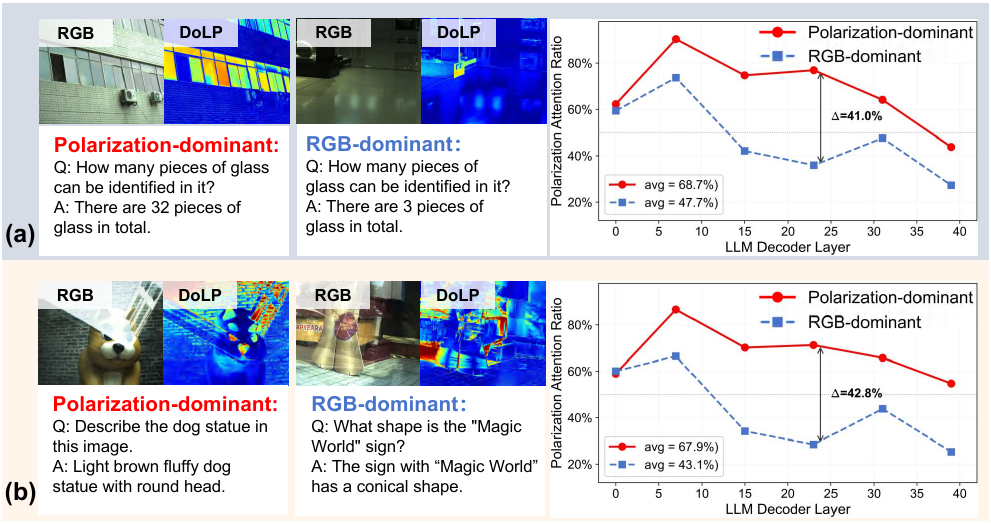}
  \caption{Adaptive attention allocation in PolarVLM. We plot the \textit{Polarization Attention Ratio} across the LLM decoder layers. The results demonstrate that PolarVLM dynamically allocates higher attention to polarimetric tokens (red) for highly ambiguous physical scenes, while relying more on RGB tokens (blue) for simpler queries, proving its capacity for context-aware modality routing without explicit fusion gates.}
  \label{fig:attention_analysis}
\end{figure}

\noindent\textbf{Adaptive attention allocation.} To gain mechanistic insight into why direct sequence concatenation outperforms explicit fusion modules, we analyze the attention mass from language tokens to visual tokens across the LLM's 40 decoder layers. Specifically, we define the \textit{Polarization Attention Ratio} as the proportion of visual attention weights allocated to polarimetric tokens versus RGB tokens. As illustrated in Fig.~\ref{fig:attention_analysis}, the LLM dynamically modulates this ratio based on the inherent optical ambiguity of the scene. For instance, in the \textit{glass counting} task (Fig.~\ref{fig:attention_analysis}(a)), a complex facade with multiple indistinguishable glass panels heavily relies on the polarization branch (68.7\% average attention). Conversely, a simpler scene with distinct visual boundaries requires significantly less polarimetric grounding (47.7\%). A similar content-aware routing emerges under reflection interference (Fig.~\ref{fig:attention_analysis}(b)): the model heavily attends to physical priors to decipher severely distorted surfaces, but reverts to RGB dominance when high-level geometric or textual cues (\eg, a printed sign) are sufficient. Layer-wise trajectories further reveal that this modality divergence peaks in the middle decoder layers, where complex semantic reasoning typically occurs. These findings compellingly demonstrate that direct concatenation empowers the LLM to function as a zero-parameter, ambiguity-driven fusion gate, adaptively gathering physical evidence only when standard RGB semantics fall short.

\section{Responsible use and asset compliance}
\label{supp:responsible_use}

PolarVQA is constructed from publicly available polarimetric datasets and does not include Internet-scraped or private personal data. All existing datasets, models, and codebases used in this work are cited, and their licenses and terms of use are respected. The released resources are intended for research on physics-aware visual reasoning, and the generated annotations are produced through structured, text-only, rule-verified prompts to reduce unsupported or unsafe content.

\end{document}